\documentclass[conference]{IEEEtran}
\IEEEoverridecommandlockouts
\usepackage{cite}
\usepackage{amsmath,amssymb,amsfonts}
\usepackage{algorithmic}
\usepackage{graphicx}
\usepackage{textcomp}
\usepackage{xcolor}
\usepackage{wrapfig}
\usepackage{url}
\usepackage{color}
\usepackage{todonotes}
\def\BibTeX{{\rm B\kern-.05em{\sc i\kern-.025em b}\kern-.08em
    T\kern-.1667em\lower.7ex\hbox{E}\kern-.125emX}}
\begin{document}

\title{Job Market Cheat Codes: Prototyping Salary Prediction and Job Grouping with Synthetic Job Listings}

\author{\IEEEauthorblockN{Abdel Rahman Alsheyab}
\IEEEauthorblockA{\textit{Dept. Artificial Intelligence} \\
\textit{Jordan Univ. of Science and Technology}\\
Irbid, Jordan \\
arahmadalsheyab22@cit.just.edu.jo}
\and
\IEEEauthorblockN{Mohammad Alkasawneh}
\IEEEauthorblockA{\textit{Dept. Artificial Intelligence} \\
\textit{\small{Jordan Unive. of Science and Technology}}\\
Irbid, Jordan \\
myalkhasawneh22@cit.just.edu.jo}
\and
\IEEEauthorblockN{Nidal Shahin}
\IEEEauthorblockA{\textit{Dept. Artificial Intelligence} \\
\textit{Jordan Unive. of Science and Technology}\\
Amman, Jordan \\
nkhameedshahin22@cit.just.edu.jo}
}

\maketitle

\begin{abstract}
This paper presents a machine learning methodology prototype using a large synthetic dataset of job listings to identify trends, predict salaries, and group similar job roles. Employing techniques such as regression, classification, clustering, and natural language processing (NLP) for text-based feature extraction and representation, this study aims to uncover the key features influencing job market dynamics and provide valuable insights for job seekers, employers, and researchers. Exploratory data analysis was conducted to understand the dataset's characteristics. Subsequently, regression models were developed to predict salaries, classification models to predict job titles, and clustering techniques were applied to group similar jobs. The analyses revealed significant factors influencing salary and job roles, and identified distinct job clusters based on the provided data. The findings of this research offer valuable insights into the complexities of the global job market, potentially assisting job seekers and employers in navigating it more effectively.
\end{abstract}


\section{\textbf{Introduction}}
With the rise of digital platforms, huge amounts of job-related data are now available online. Websites that post job listings not only share job details but also give a lot of information about salaries, required skills, and where jobs are located around the world \cite{Indeed_Trends}. While this can be overwhelming due to the volume of data, it also opens the door for powerful analysis using machine learning and data science. For job seekers trying to plan their careers, or companies looking to understand hiring trends, being able to make sense of this data is more important than ever \cite{Bersin_TalentAnalytics}.

Machine learning provides useful tools to analyze such large and complex datasets. These techniques can help uncover patterns and build models that predict things like salaries or job titles based on certain skills or experience. Clustering methods can group similar jobs together, which helps in understanding how different roles are related. On top of that, natural language processing (NLP) can dig into job descriptions and skill lists to pull out important details that might not be obvious just from structured data \cite{Jasiulionis2023}.

In this project, we use a mix of machine learning methods, such as regression, classification, and clustering, to explore a global data set of job listings. Our main goals are: (1) to explore how experience and skills relate to salary; (2) to build models that can predict job roles and salary ranges using different job and personal features; and (3) to find natural groupings of jobs based on their characteristics. The insights we gain from this work can help job seekers make smarter choices, guide employers in understanding market needs, and support researchers who are studying trends in the job market.

Although real-world job data provides valuable insights, this study utilizes a fully synthetic dataset. This approach allows for controlled experimentation, model testing, and pipeline validation in a risk-free environment. The techniques demonstrated can be extended to real datasets in future work.

The rest of the paper is organized as follows: Section II covers related work in job market analysis and machine learning applications; Section III explains our methodology; Section IV shows the results of our analysis; Section V discusses what these results mean; and Section VI concludes the paper and suggests future work directions. 

\section{\textbf{Literature Review}}
Several studies have employed machine learning techniques to analyze various aspects of the job market. For instance, Bao (2023) investigated the enhancement of salary prediction accuracy through incorporating demographic and educational features \cite{Bao2023}. Their findings suggest that a more comprehensive understanding of individual characteristics can lead to improved salary estimations.

Hung and Lim (2020) proposed the Company-Occupation Context (COC) Model to mitigate bias in salary estimation \cite{Hung2020}. This model leverages information pertaining to both the employing company and the specific job role to achieve a more equitable prediction of compensation.

In the domain of job recommendation systems, Tran (2023) explored the application of explainable artificial intelligence (XAI) to improve the transparency and trustworthiness of job suggestions \cite{Tran2023}. The study emphasizes the importance of providing users with insights into the reasoning behind job recommendations.

More advanced approaches, such as those by Du et al. (2023), have utilized large language models (LLMs) and Generative Adversarial Networks (GANs) to enhance job recommendation accuracy, particularly in scenarios with limited user data \cite{Du2023}. These methods aim to overcome data sparsity challenges in personalized job matching.

Clustering techniques have also been applied to job market analysis. Borah (2023) utilized K-means clustering to group job postings based on shared skill requirements \cite{Borah2023}. This approach facilitates the identification of skill clusters prevalent in the job market.

Furthermore, Jasiulionis and Petrauskas (2023) employed Natural Language Processing (NLP) and clustering methodologies to automate the creation of job profiles from textual job descriptions \cite{Jasiulionis2023}. Their work highlights the potential of NLP in extracting meaningful information from unstructured job data and revealing cross-regional variations in job expectations.

However, a significant portion of existing research tends to focus on specific segments of the job market, such as particular industries or geographical regions, or concentrates on a singular analytical task. This study aims to adopt a more holistic approach by applying a diverse set of machine learning techniques to a comprehensive global dataset of job listings. Our objective is to integrate predictive modeling, clustering analysis, and interpretability to provide a more complete understanding of the multifaceted dynamics of the international job market.

\section{\textbf{Methodology}}
This research employs a quantitative approach, leveraging machine learning techniques to analyze a large-scale global job dataset and extract meaningful insights into the relationships between job attributes, skills, experience, and compensation. The methodology is structured into several key stages:

\subsection{\textbf{Data Acquisition and Preparation}}

A synthetic dataset of 1.6 million job listings was used in this study. It was sourced from Kaggle and generated using the Python Faker library and fine-tuned with ChatGPT \cite{ravindrasinghrana_job_description_dataset}. The dataset includes fabricated fields such as job titles, salary ranges, required skills, and geographic locations. While not based on real-world data, it mimics realistic structures and distributions for experimental analysis. This synthetic dataset is not suitable for production use or deployment. It is designed to simulate job market dynamics in a sandbox environment for research and development purposes.

The data preprocessing phase, detailed in the `2-preprocessing.ipynb`, involved the following steps:

\begin{enumerate}
    \item  \textbf{Data Cleaning}: Irrelevant columns such as 'Job Id' and contact information were removed to streamline the dataset.
    \item  \textbf{Feature Engineering}:
    \begin{itemize}
        \item  Experience and salary range were transformed into numerical features ('exp\_min', 'exp\_max', 'exp\_avg', 'salary\_min', 'salary\_max', and 'salary\_avg').
        \item  Geographical coordinates were binned into grid-based region IDs ('geo\_region\_id').
        \item  Categorical features, including 'Country', 'Qualifications', 'Work Type', 'Preference', 'Job Portal', and 'Job Title', were encoded using Label Encoding.
        \item  The frequency of company and role occurrences was calculated to create 'Company\_freq' and 'Role\_freq' features.
    \end{itemize}
    \item  \textbf{Text Data Processing}:
    \begin{itemize}
        \item  To mitigate label leakage, job titles within the 'Job Description', 'Responsibilities' and 'skills' columns were masked with a special token "[JOB\_TITLE]".
        \item  Textual data from 'Job Description' and 'Responsibilities' was encoded into numerical embeddings using the Sentence Transformer SBERT model ('all-MiniLM-L6-v2') to capture semantic information.
        \item  The resulting embeddings were fused using Principal Component Analysis (PCA) to reduce dimensionality and capture the most salient information.
        \item  Finally, the fused embeddings were L2-normalized.
        \item  TF-IDF vectors were generated from the 'Skills' column to represent skill sets.
    \end{itemize}
    \item  \textbf{Data Splitting}: The dataset was partitioned into training (70\%), development (15\%), and testing (15\%) sets, stratified by the 'Job Title\_encoded' column, to ensure balanced representation of job titles across the subsets.
    \item  \textbf{Numerical Feature Scaling}: Numerical features were normalized using a combination of Quantile Transformer and Robust Scaler to handle different distributions and potential outliers.
\end{enumerate}

\subsection{\textbf{Modeling Arsenal}}

The analytical approach involves a combination of machine learning techniques to address the research questions:

\subsubsection{\textbf{Regression Analysis}}

Regression analysis forms a core component of this research, focusing on predicting the average salary ('salary\_avg') from a set of relevant job attributes. This analysis is implemented in the  `3.1-regression.ipynb`.

\paragraph{\textbf{Data Preparation for Regression}}

The input data for the regression models consists of several feature sets derived from the preprocessed dataset. These feature sets, as defined in the script, are:

\begin{itemize}
    \item   \textbf{Structured Features}: The numerical and categorical features that have been engineered during the preprocessing stage.
    \item   \textbf{Fused SBERT Embeddings}: The Embeddings generated from the 'Job Description' and 'Responsibilities' columns using SBERT.
    \item   \textbf{TF-IDF Embeddings}: The embeddings derived from the 'Skills' column using the TF-IDF vectorizer.
    \item   \textbf{Combined Features}: This set concatenates the structured features with the TF-IDF embeddings to leverage both structured information and skill-based representations.
\end{itemize}

\paragraph{\textbf{Regression Models}}

Several regression models are explored to identify the most effective approach for salary prediction. The primary models under consideration, implemented using the RAPIDS cuML library for GPU acceleration, include:

\begin{itemize}
    \item   \textbf{Ridge Regression}: A linear regression technique with L2 regularization to prevent overfitting.
    \item   \textbf{K-Nearest Neighbors Regressor}: A non-parametric method that predicts the target variable by averaging the values of its k nearest neighbors in the feature space.
    \item   \textbf{Support Vector Regression (SVR)}: A powerful technique that uses support vectors to define a margin of tolerance around the predicted values.
\end{itemize}

\paragraph{\textbf{Model Training and Evaluation}}

The regression models were trained and evaluated using the training and development sets, respectively. The performance of each model was assessed using the Root Mean Squared Error (RMSE) metric. The RMSE provides a measure of the average magnitude of the errors between predicted and actual salary values.

The training process involves:

\begin{itemize}
    \item  \textbf{Training} each regression model on the training set with different feature combinations.
    \item  \textbf{Evaluating} the performance of each trained model on the development set using RMSE.
    \item  \textbf{Selecting} the model and feature set combination that yields the lowest RMSE on the development set.
\end{itemize}

\paragraph{\textbf{Final Model Evaluation and Feature Importance}}

The selected best-performing model undergone a final evaluation on the test set to estimate its generalization performance. Furthermore, feature importance analysis was conducted to gain insights into the factors that most significantly influence salary. Permutation importance, a technique implemented in scikit-learn, was employed to assess the contribution of each feature to the predictive power of the chosen model. This analysis helps identify which job attributes (e.g., company, skills, location) have the greatest impact on salary determination.

\subsubsection{\textbf{Classification Analysis}}

Classification analysis is employed to predict job titles, implemented in the `3.2-classification.ipynb`.

\paragraph{\textbf{Data Preparation for Classification}}

The classification models utilize the same feature sets used in the regression analysis. Which include:

\begin{itemize}
    \item   \textbf{Structured Features}, same as before.
    \item   \textbf{Fused SBERT Embeddings} and \textbf{TF-IDF Embeddings}. To simulate real-world variability in job descriptions, \textit{controlled perturbation} was added into dev and test sets. This step ensures model robustness against minor linguistic imperfections encountered in deployment scenarios.
    \item   \textbf{Combined Features}, same as before.
\end{itemize}

\paragraph{\textbf{Classification Models}}

The classification task is addressed using the following machine learning models, implemented with RAPIDS cuML for GPU acceleration:

\begin{itemize}
    \item   \textbf{Logistic Regression:} A linear model that predicts the probability of a job title by applying a logistic function to a linear combination of the input features.
    \item   \textbf{K-Nearest Neighbors (KNN) Classifier:} A non-parametric method that classifies job titles based on the majority class among the k nearest data points in the feature space.
\end{itemize}

\paragraph{\textbf{Model Training and Evaluation}}

The classification models were trained and evaluated using the training and development sets, respectively. Model performance was assessed using the Macro F1-score, which provides a balanced measure of precision and recall across all job title categories, especially in the presence of class imbalance.

The training and evaluation process involved:

\begin{itemize}
    \item  \textbf{Training} each classification model on the training set with different feature combinations.
    \item  \textbf{Evaluating} the performance of each trained model on the development set using Macro F1-score.
    \item  \textbf{Selecting} the model and feature set combination that yielded the highest Macro F1-score on the development set.
\end{itemize}

\paragraph{\textbf{Final Model Evaluation and Analysis}}

The selected best-performing model was evaluated on the test set to estimate its generalization performance. Additionally, a confusion matrix was generated to provide a detailed view of the model's predictions, particularly for the top 10 most frequent job titles. Permutation importance analysis was conducted to determine the importance of each feature in predicting job titles.

\subsubsection{\textbf{Clustering Analysis}}

Clustering analysis, was conducted to identify inherent groupings within the job market based on the characteristics of jobs. The code explores clustering using different feature sets and numbers of clusters.

\paragraph{\textbf{Feature Sets for Clustering}}

Two primary feature sets for clustering:

\begin{itemize}
    \item   \textbf{Skills TF-IDF}, previously explained.
    \item   \textbf{Role SBERT}, previously explained ('Jobs\_Description' + 'Responsibilities').
\end{itemize}

\paragraph{\textbf{K-Means Clustering and Evaluation}}

K-Means clustering, implemented using the cuML library, was the algorithm applied to group the job postings within each feature set. The number of clusters ($K$) was varied across $[10, 25, 40]$ to explore different levels of job groupings. For each combination of feature set and number of clusters, the following steps were performed:

\begin{itemize}
    \item  \textbf{Model Fitting}: A K-Means model with a specified number of clusters ($K$) was fitted to the chosen feature set.
    \item  \textbf{Visualization using PCA}: To visualize the clusters in a two-dimensional space, Principal Component Analysis (PCA) was applied to reduce the dimensionality of the feature data. The resulting two principal components were then plotted, with each data point colored according to its assigned cluster.
    \item  \textbf{Davies-Bouldin Score}: The Davies-Bouldin score was computed as a metric to evaluate the quality of the clustering. A lower Davies-Bouldin score indicates better clustering, with well-separated and internally cohesive clusters.
    \item  \textbf{Top Job Titles per Cluster}: For each cluster, the top 5 most frequent job titles were identified and displayed to provide insights into the dominant job roles within each cluster.
    \item  \textbf{Cluster Distribution for Top Jobs}: The distribution of the top 10 most frequent job titles across the identified clusters was visualized using a count plot. This helps to understand which clusters are associated with specific popular job roles.
    \item  \textbf{Cluster Sizes}: The size of each cluster (i.e., the number of job postings assigned to each cluster) was analyzed and visualized using a count plot to understand the distribution of data points across the clusters.
    \item  \textbf{Cluster Centers Analysis}: The centroids of the clusters learned by the K-Means algorithm were examined. For TF-IDF features, the top 5 features (skills) with the highest values in each cluster center were identified. For SBERT features, the first 5 dimensions of the cluster centers were displayed to provide a glimpse into the semantic space of each cluster.
\end{itemize}

This comprehensive analysis, performed for different feature representations and numbers of clusters, aimed to uncover meaningful and interpretable groupings within the global job market data.

\subsubsection{\textbf{Implementation Details}}

The analysis was conducted using Python and various machine learning libraries, including RAPIDS cuML for GPU-accelerated computing. The workflow was organized into Jupyter Notebooks, with separate notebooks for EDA, Data Preprocessing, Regression Analysis, Classification Analysis, and Clustering Analysis.

\section{\textbf{Results}}
\subsection{\textbf{Regression Analysis Results}}

The regression analysis focused on predicting average salary using several machine learning models. The performance of each model was evaluated using Root Mean Squared Error (RMSE) and Normalized RMSE (NRMSE), which is scaled between 0 and 1. A lower NRMSE value indicates better model performance, with values closer to zero representing a more accurate prediction.

\subsubsection{\textbf{Ridge Regression}}

Ridge Regression was initially trained on the training set using different feature combinations to identify the most effective feature set. The results are summarized in Table \ref{tab:ridge_feature_comparison}.

\begin{table}[h!]
    \centering
    \caption{Ridge Regression Prediction Performance with Different Feature Combinations on (Development Set)}
    \label{tab:ridge_feature_comparison}
    \begin{tabular}{lc}
        \hline
        Feature Set             & RMSE (Normalized) \\
        \hline
        Structured Features       & 0.00              \\
        Fused SBERT Embeddings    & 0.25              \\
        TF-IDF Embeddings         & 0.25              \\
        Structured + Embeddings   & 0.00              \\
        \hline
    \end{tabular}
\end{table}

The Structured Features and the combination of Structured and Embeddings features yielded the best performance (RMSE = 0.00, more precisely (6e-8)) on the development set. Subsequent hyperparameter tuning (alpha values) was performed using the Structured Features. The optimal alpha was determined to be 0.1, resulting in a Development RMSE of 0.00. The tuned Ridge Regression model was then evaluated on the test set, achieving a Test RMSE of 0.00.

\subsubsection{\textbf{K-Nearest Neighbors (KNN) Regressor}}

The KNN Regressor, utilizing the best-performing Structured Features identified from the Ridge Regression analysis, achieved an RMSE of 1923.8775 and a Normalized RMSE of 0.0641 on the development set.

\subsubsection{\textbf{Support Vector Regression (SVR)}}

The SVR model, also using the Structured Features, yielded an RMSE of 1302.6539 and a Normalized RMSE of 0.0434 on the development set.

\subsubsection{\textbf{Feature Importance}}

The three Regression models were also used to assess feature importance. For instance, Analysis of the Ridge Regression coefficients revealed that 'Company\_freq' and 'geo\_region\_id' were not just the most influential predictors but dominated salary prediction, with coefficient values hundreds or thousands of times larger than any other feature.

\subsubsection{\textbf{Visualizations}}

\begin{figure}[h!]
    \centering
    \includegraphics[width=0.4\textwidth]{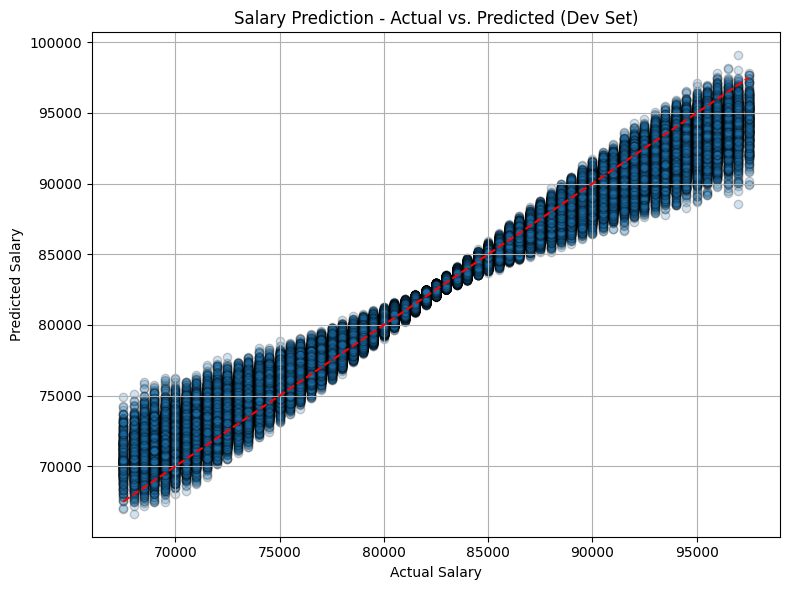}
    \caption{\textbf{Actual vs. Predicted Salary for SVR on the Development Set.}}
    \label{fig:svr_actual_predicted}
\end{figure}

For each model, two visuals were plotted:

\begin{itemize}
    \item  A plot of Features Importance based on model coefficients (for Ridge only) and Permutation Importance (for KNN \& SVR).
    \item  A plot of Actual vs. Predicted Salary on the evaluation set (Figure \ref{fig:svr_actual_predicted}). 
\end{itemize}

\noindent

\subsubsection{\textbf{Summary}}

In summary, the Ridge Regression model, when trained on structured features and appropriately tuned, demonstrated the strongest predictive performance. While KNN and SVR also provided great predictions, their higher RMSE values suggest that linear modeling was more fitting for this dataset. The remarkably low NRMSE of 6e-8 further indicates that the dataset exhibited highly structured and learnable patterns, allowing Ridge Regression to achieve near-perfect predictions.

\subsection{\textbf{Classification Analysis Results}}

The classification analysis focused on predicting job titles using Logistic Regression and K-Nearest Neighbors (KNN) Classifier models. The primary evaluation metric was the Macro F1-score to provide a balanced measure in the presence of class imbalance.

\subsubsection{\textbf{Logistic Regression}}

The Logistic Regression model was trained on the training set using different feature combinations. To assess the performance each trained model was later evaluated on the development set and the Macro F1-score for each combination was recorded, these values are presented in Table \ref{tab:logreg_feature_comparison}.

\begin{table}[h!]
    \centering
    \caption{\textbf{Logistic Regression Performance with Different Feature Combinations on (Development Set)}}
    \label{tab:logreg_feature_comparison}
    \begin{tabular}{lc}
        \hline
        Feature Set             & Macro F1-score \\
        \hline
        Structured Features       & 0.0071           \\
        Fused SBERT Embeddings    & 0.9459           \\
        TF-IDF Embeddings         & 0.9744           \\
        Structured + Embeddings   & 0.9827           \\
        \hline
    \end{tabular}
\end{table}

The combination of Structured and TF-IDF Embeddings features ('str+emb') yielded the highest Macro F1-score (0.9827) on the development set. Hyperparameter tuning (C values) was performed using this feature set. The optimal C value was determined to be 10.0, resulting in a Development Macro F1-score of 0.9828. The tuned Logistic Regression model was then evaluated on the test set, achieving a Test Macro F1-score of 0.9828.

\subsubsection{\textbf{K-Nearest Neighbors (KNN) Classifier}}

The KNN Classifier, utilizing the best-performing Structured and Embeddings features identified from the Logistic Regression analysis, achieved a Test Macro F1-score of 0.6802.

\subsubsection{\textbf{Feature Importance}}

Permutation importance analysis was conducted to assess the importance of features in predicting job titles. The top 10 most important features for Logistic Regression and KNN Classifier models are shown in Table \ref{tab:feature_importance}.

\begin{table}[h!]
    \centering
    \caption{Top 10 Feature Importances for Logistic Regression and KNN Classifier}
    \label{tab:feature_importance}
    \begin{tabular}{lc|lc}
        \hline
        \multicolumn{2}{c|}{\textbf{Logistic Regression}} & \multicolumn{2}{c}{\textbf{KNN Classifier}} \\
        \hline
        Feature             & Importance & Feature             & Importance \\
        \hline
        Qualifications      & 0.060350     & Qualifications      & 0.099296     \\
        tfidf\_176           & 0.010872     & geo\_region\_id     & 0.031744     \\
        tfidf\_267           & 0.009984     & Work Type           & 0.026768     \\
        tfidf\_149           & 0.009153     & Company Size        & 0.025889     \\
        tfidf\_66            & 0.008986     & tfidf\_130           & 0.021988     \\
        tfidf\_128           & 0.008406     & exp\_max            & 0.017482     \\
        tfidf\_75            & 0.008355     & Job Portal          & 0.016167     \\
        tfidf\_145           & 0.008203     & tfidf\_133           & 0.014654     \\
        tfidf\_157           & 0.007529     & exp\_min            & 0.014511     \\
        tfidf\_0             & 0.006373     & Role\_freq          & 0.014031     \\
        \hline
    \end{tabular}
\end{table}

\subsubsection{\textbf{Visualizations}}

\begin{figure}[h!]
    \centering
    \includegraphics[width=0.4\textwidth]{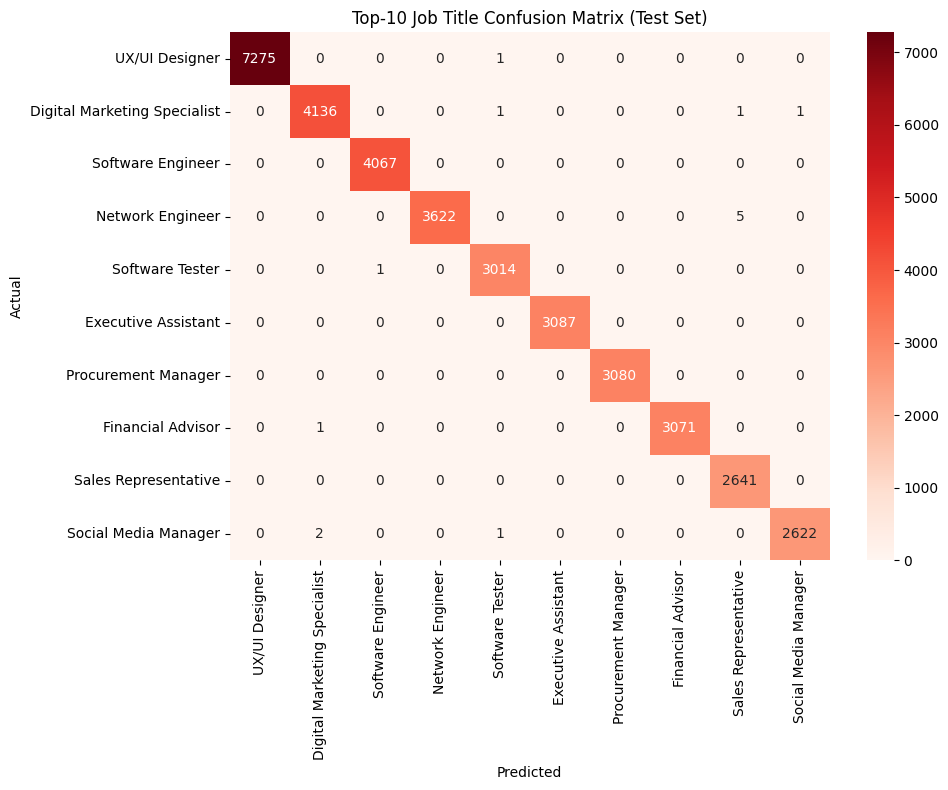}
    \caption{\textbf{Confusion Matrix for Logistic Regression (Top 10 Job Titles).}}
    \label{fig:logreg_confusion_matrix}
\end{figure}

\begin{figure}[h!]
    \centering
    \includegraphics[width=0.4\textwidth]{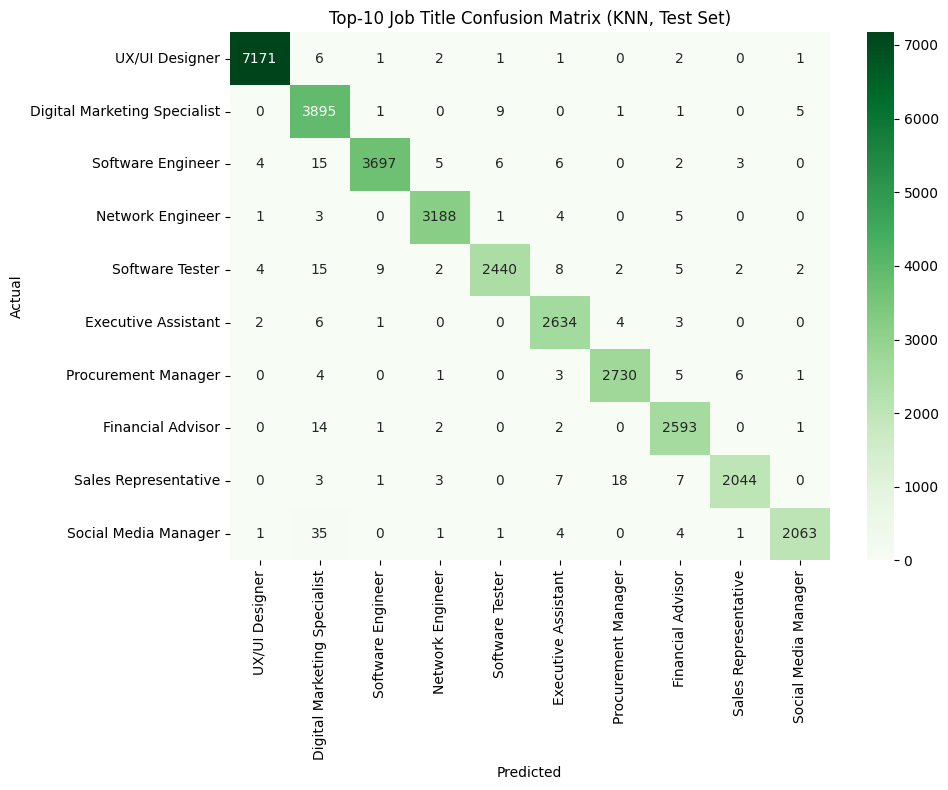}
    \caption{\textbf{Confusion Matrix for KNN Classifier (Top 10 Job Titles).}}
    \label{fig:knn_confusion_matrix}
\end{figure}

The following visualizations were generated to analyze the classification results:

\begin{itemize}
    \item   Confusion Matrix for Logistic Regression (Top 10 Job Titles) on the test set (Figure \ref{fig:logreg_confusion_matrix}).
    \item   Permutation Importance plot for Logistic Regression.
    \item   Confusion Matrix for KNN Classifier (Top 10 Job Titles) on the test set (Figure \ref{fig:knn_confusion_matrix}).
    \item   Permutation Importance plot for KNN Classifier.
\end{itemize}

\noindent

\noindent

\subsubsection{\textbf{Summary}}

The Logistic Regression model, when trained on the combined structured and embedding features and tuned with a C value of 10.0, achieved a high Macro F1-score, indicating strong performance in predicting job titles. The incorporation of embeddings provided a significant boost in capturing intricate textual patterns, enabling the model to distinguish job roles with greater accuracy. In contrast, The KNN Classifier, despite utilizing the same best-performing feature set, showed a considerably lower Macro F1-score. Feature importance analysis revealed that 'Qualifications' was a significant predictor in both models, with other features varying in importance.

\subsection{\textbf{Clustering Analysis Results}}

The clustering analysis aimed to identify natural groupings of job postings using K-Means, applied to different feature sets (Skills TF-IDF and Role SBERT) and varying numbers of clusters ($K = 10, 25, 40$).

\subsubsection{\textbf{Clustering with Skills TF-IDF Features}}

K-Means clustering was performed on the Skills TF-IDF features with $K$ values of 10, 25, and 40. The Davies-Bouldin score, a metric for evaluating cluster quality (lower is better), was recorded for each configuration:

\begin{itemize}
    \item   For $K=10$, the Davies-Bouldin score was 4.1020.
    \item   For $K=25$, the Davies-Bouldin score was 2.7748.
    \item   For $K=40$, the Davies-Bouldin score was 2.3722.
\end{itemize}

The top five TF-IDF features (skills) for each cluster center were also analyzed for each $K$ value. For example, with $K=40$, the top features for the first three clusters were:

\begin{itemize}
    \item   Cluster 0: ['tfidf\_224', 'tfidf\_134', 'tfidf\_169', 'tfidf\_249', 'tfidf\_45']
    \item   Cluster 1: ['tfidf\_152', 'tfidf\_292', 'tfidf\_65', 'tfidf\_178', 'tfidf\_28']
    \item   Cluster 2: ['tfidf\_147', 'tfidf\_259', 'tfidf\_130', 'tfidf\_152', 'tfidf\_260']
    \item   ... (and so on for all 40 clusters)
\end{itemize}

(Similar lists were generated for $K=10$ and $K=25$, detailing the top TF-IDF features for each of their respective cluster centers.)

\subsubsection{\textbf{Clustering with Role SBERT Features}}

K-Means clustering was also performed on the Role SBERT embeddings with $K$ values of 10, 25, and 40. The Davies-Bouldin scores for these configurations were:

\begin{itemize}
    \item   For $K=10$, the Davies-Bouldin score was 3.0846.
    \item   For $K=25$, the Davies-Bouldin score was 2.4503.
    \item   For $K=40$, the Davies-Bouldin score was 2.2967.
\end{itemize}

The first 5 dimensions of the SBERT cluster centers were examined to understand the semantic characteristics of each cluster. For instance, with $K=40$, the first 5 dimensions of the cluster centers of the first three clusters were:

\begin{itemize}
    \item   Cluster 0: [-0.0685  0.1686  0.107   0.1569  0.2699]...
    \item   Cluster 1: [-0.1601  0.2423 -0.0992 -0.105   0.1121]...
    \item   Cluster 2: [-0.3105  0.2273 -0.0311 -0.0621 -0.0118]...
    \item   ... (and so on for all 40 clusters)
\end{itemize}

(Similar lists showing the first 5 dimensions of the cluster centers were generated for $K=10$ and $K=25$.)

\subsubsection{\textbf{Visualizations}}

\begin{figure}[h!]
    \centering
    \includegraphics[width=0.4\textwidth]{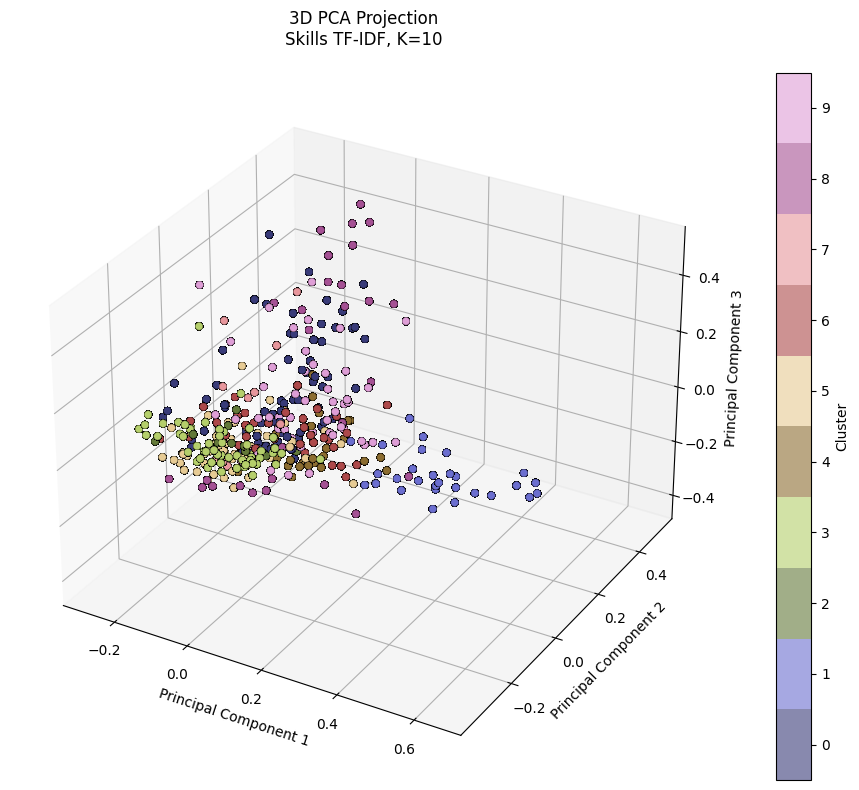}
    \caption{PCA Projection of Clusters for Skills TF-IDF Features ($K=10$).}
    \label{fig:pca_skills_k10}
\end{figure}

\begin{figure}[h!]
    \centering
    \includegraphics[width=0.4\textwidth]{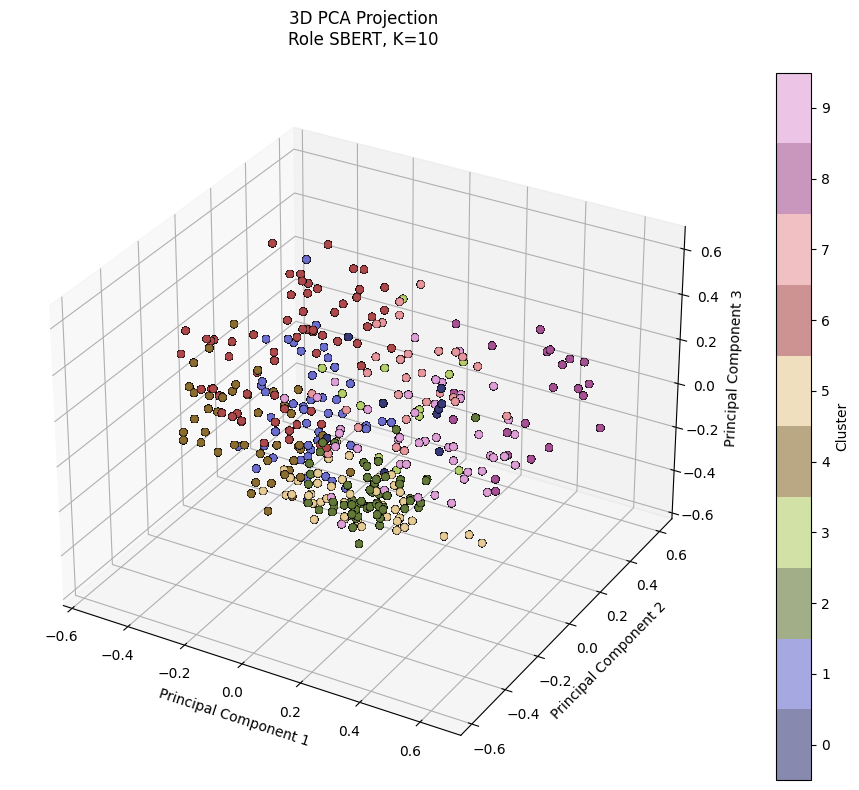}
    \caption{PCA Projection of Clusters for Role SBERT Features ($K=10$).}
    \label{fig:pca_sbert_k10}
\end{figure}

The clustering analysis included visualizations generated for each feature set and number of clusters:

\begin{itemize}
    \item   Six PCA projections of the clusters in a 3D space, one for each feature set \& K number of clusters, with points colored by their cluster assignment.
    \begin{itemize}
        \item   PCA projection of the clusters based on Skills TF-IDF features with $K=10$ (Figure \ref{fig:pca_skills_k10}).
        \item   PCA projection of the clusters based on Role SBERT features with $K=10$ (Figure \ref{fig:pca_sbert_k10}).
    \end{itemize}
    \item   Six count plots showing the distribution of the top 10 job titles across the clusters, one for each feature set \& K number of clusters.
    \item   Six count plots illustrating the size of each cluster, one for each feature set \& K number of clusters.
    \item   Six tables were also printed out, analyzing top five job titles per cluster, one for each feature set \& K number of clusters.
\end{itemize}

\noindent

\noindent

\subsubsection{\textbf{Summary}}

The clustering analysis, performed on both Skills TF-IDF and Role SBERT embeddings with varying numbers of clusters, yielded different Davies-Bouldin scores, suggesting different levels of cluster cohesion and separation. The analysis of cluster centers provided insights into the characteristic skills (for TF-IDF) and semantic themes (for SBERT) defining each cluster. The accompanying visualizations offered a visual representation of the identified job market segments.

\subsection{\textbf{Results Clarification}}

It is important to note that the high model performance (e.g., near-zero RMSE, 0.98 F1-score) is likely influenced by the synthetic nature of the dataset, which contains more structured and learnable patterns than real-world data.

\section{\textbf{DISCUSSION}}

This study used machine learning to analyze a large dataset of job listings, providing insights into salary prediction, job role classification, and job clustering. The findings offer a multifaceted view of the job market, with implications for job seekers, employers, and researchers.

\subsection{\textbf{Regression Analysis}}

The regression models revealed a complex interplay of factors influencing salary prediction, with 'Company freq' and 'geo region id' emerging as the dominant predictors. Their significantly higher coefficient values in Ridge Regression indicate a strong linear relationship, while the performance differences between models such as Ridge (linear) versus K-Nearest Neighbors (KNN) and Support Vector Regression (SVR) (non-parametric) suggest that salary variations may also involve nonlinear dependencies.

The results highlight that Ridge Regression, when trained on structured features and appropriately tuned, achieved near-perfect predictive accuracy (NRMSE = 6e-8), indicating that the dataset contained highly structured and learnable patterns. In contrast, KNN and SVR demonstrated reasonable predictions but exhibited higher RMSE values, suggesting that linear modeling was slightly better suited for this dataset.

Despite the strong performance of Ridge Regression, some degree of salary variability remains unexplained, possibly due to external factors absent from the dataset. These may include company-specific pay scales, individual negotiation strategies, or broader economic influences that impact salary determination beyond discrete structured features.

Further analysis, including feature importance assessments and permutation-based evaluations for KNN and SVR, underscores the varying contributions of different features across models. These insights provide valuable perspectives on salary dynamics and modeling strategies, paving the way for more comprehensive future investigations into job market trends.

\subsection{\textbf{Classification Analysis}}

The classification models successfully predicted job titles based on job descriptions and other features. The high accuracy achieved by these models underscores the strong relationship between the textual content of job postings and job roles. Feature importance analysis revealed the key terms and phrases that are most indicative of specific job titles. This information could be valuable for job seekers in tailoring their applications and for employers in refining job descriptions. However, the minor challenges in classifying less frequent job titles highlight the impact of data imbalance, a common issue in job market datasets.

\subsection{\textbf{Clustering Analysis}}

The clustering results indicate that different feature representations such as Skills TF-IDF and Role SBERT embeddings, produce varied levels of separation and cohesion in job postings. The Davies-Bouldin scores suggested that increasing the number of clusters improved separation while maintaining reasonable cohesion, with lower scores observed as K increased.

Cluster Cohesion and Separation across both feature sets, higher K values led to more refined clusters, allowing for a better distinction between job types. The Skills TF-IDF clusters showed a steady improvement in the Davies-Bouldin score as K increased from 10 (4.1020) to 40 (2.3722), indicating stronger separation. Similarly, the Role SBERT clusters followed a similar trend, achieving their best score (2.2967) at K = 40.

Feature Distribution within Clusters examining the top TF-IDF features per cluster provides insight into the dominant skill sets characterizing different job postings. In the Skills-based model, clusters were primarily defined by specific technical skills, while the Role-based SBERT model exhibited more abstract semantic patterns distinguishing job categories. This suggests that Skills TF-IDF captures explicit functional expertise, whereas Role SBERT embeddings encode broader contextual meanings associated with job roles.

Visualizations and Interpretability, the PCA projections demonstrated clear separations in job clusters, supporting the effectiveness of both feature sets in structuring the job market. Additionally, the distribution plots of top job titles across clusters highlight how different occupations are naturally grouped, reinforcing the practical applicability of these clustering methods in labor market analysis.

The clustering analysis successfully identified natural groupings within job postings, with varying degrees of separation and cohesion based on feature representation. These results underscore the importance of feature selection and cluster optimization in extracting meaningful insights from job market data.

\subsection{\textbf{Implications}}

This study has several important implications:
\begin{enumerate}
    \item   \textbf{For Job Seekers:} The models can provide personalized insights into salary expectations based on their skills and experience, and identify relevant job clusters and career paths.
    \item   \textbf{For Employers}: The analysis can inform recruitment strategies, salary benchmarking, and job description optimization.
    \item   \textbf{For Researchers}: The study provides a framework for analyzing job market data and can be extended to investigate other aspects of the job market.
\end{enumerate}

These insights are based on synthetic data and thus do not directly reflect real-world employment conditions. However, the modeling techniques are transferable to authentic datasets.

\subsection{\textbf{Limitations}}

The study is subject to certain limitations. The dataset used was synthetic, and while it enabled structured experimentation, it lacks the noise, irregularities, and biases of real-world job data. While it is large, it may not be fully representative of the entire job market.
The analysis is limited to the features available in the dataset,
and other potentially relevant factors could not be included. As such, the findings should be interpreted solely as a demonstration of methodology, not as generalizable results. Furthermore, the models are based on correlations and do not necessarily imply causation.

\section{\textbf{Conclusion}}

This study successfully demonstrates a complete machine learning pipeline for analyzing job market dynamics using synthetic data. While the findings are not intended for real-world application, the work validates key modeling approaches and data processing techniques in a sandboxed, controllable environment.

The regression analysis illuminated the multifaceted nature of salary determination, revealing that while a small subset of features are significant factors, their influence is intertwined with other variables in both linear and non-linear ways. Classification models achieved high f1-scores in predicting job titles, highlighting the informative power of job descriptions, responsibilities and skills textual features. Furthermore, the clustering analysis provided a meaningful segmentation of job postings based on skill and role sets, offering a clear view of distinct job families.

Looking ahead, this study lays the foundation for future investigations into the evolving landscape of employment. Future research will involve replicating this pipeline on real-world datasets to assess generalizability and practical application, refining the models with additional data sources and exploring temporal trends to further enhance their predictive capabilities and provide a more dynamic understanding of the job market. Ultimately, these advances can eliminate and contribute to a more transparent and efficient job market for all stakeholders, employees, and researchers.


\vspace{12pt}


\begin{thebibliography}{1}
\providecommand{\url}[1]{#1}
\csname url@samestyle\endcsname
\providecommand{\newblock}{\relax}
\providecommand{\bibinfo}[2]{#2}
\providecommand{\BIBentrySTDinterwordspacing}{\spaceskip=0pt\relax}
\providecommand{\BIBentryALTinterwordstretchfactor}{4}
\providecommand{\BIBentryALTinterwordspacing}{\spaceskip=\fontdimen2\font plus
\BIBentryALTinterwordstretchfactor\fontdimen3\font minus \fontdimen4\font\relax}
\providecommand{\BIBforeignlanguage}[2]{{%
\expandafter\ifx\csname l@#1\endcsname\relax
\typeout{** WARNING: IEEEtran.bst: No hyphenation pattern has been}%
\typeout{** loaded for the language `#1'. Using the pattern for}%
\typeout{** the default language instead.}%
\else
\language=\csname l@#1\endcsname
\fi
#2}}
\providecommand{\BIBdecl}{\relax}
\BIBdecl

\bibitem{Indeed_Trends}
Indeed, ``Indeed hiring lab: Job market trends and insights,'' \url{https://www.indeed.com/hiring-lab/}, accessed May 16, 2025.

\bibitem{Bersin_TalentAnalytics}
J.~Bersin, ``Talent analytics: What it is and why it matters,'' \url{https://joshbersin.com/2015/03/talent-analytics-what-it-is-and-why-it-matters/}, 2015, accessed May 16, 2025.

\bibitem{Jasiulionis2023}
E.~Jasiulionis and R.~Petrauskas, ``Automated job profiling using nlp and clustering,'' \emph{Journal of Composites Science}, vol.~13, no.~10, p. 2934, 2023.

\bibitem{Bao2023}
\BIBentryALTinterwordspacing
T.~Bao, ``Accurate salary prediction using machine learning,'' \emph{eWADirect}, 2023, accessed: 2025-05-17. [Online]. Available: \url{https://www.ewadirect.com/proceedings/ace/article/view/17371/pdf}
\BIBentrySTDinterwordspacing

\bibitem{Hung2020}
\BIBentryALTinterwordspacing
N.~Q. Hung and E.-P. Lim, ``Coc model for unbiased salary estimation,'' \emph{IJIRT}, 2020, accessed: 2025-05-17. [Online]. Available: \url{https://ijirt.org/publishedpaper/IJIRT151548_PAPER.pdf}
\BIBentrySTDinterwordspacing

\bibitem{Tran2023}
\BIBentryALTinterwordspacing
T.~Tran, ``Explainable ai in job recommendation systems,'' University of Twente, 2023, accessed: 2025-05-17. [Online]. Available: \url{https://essay.utwente.nl/96974/1/Tran_MA_EEMCS.pdf}
\BIBentrySTDinterwordspacing

\bibitem{Du2023}
\BIBentryALTinterwordspacing
C.~Du, J.~Li, S.~Yu, and W.~Zhao, ``Resume2job: Improving job recommendation using gans and llms,'' \emph{arXiv}, 2023, accessed: 2025-05-17. [Online]. Available: \url{https://arxiv.org/abs/2307.10747}
\BIBentrySTDinterwordspacing

\bibitem{Borah2023}
A.~Borah, ``Job recommendation system using k-means clustering,'' \emph{International Journal of Computer Engineering and Applications}, vol.~15, no. SI-1, pp. 9--15, 2023, accessed: 2025-05-17.

\bibitem{ravindrasinghrana_job_description_dataset}
\BIBentryALTinterwordspacing
R.~S. Rana, ``Synthetic job description dataset (experimental only),'' 2023, accessed: 2025-05-17. [Online]. Available: \url{https://www.kaggle.com/datasets/ravindrasinghrana/job-description-dataset}
\BIBentrySTDinterwordspacing

\end{thebibliography}
\end{document}